\documentclass[twoside]{article}
\usepackage{amsfonts,amssymb,amsbsy,textcomp,marvosym,picins,amsmath,caption,threeparttable,amsthm,subfigure,float,lastpage,lscape}
\usepackage{eurosym,mathrsfs,fancyhdr,CJK,multicol,graphics,indentfirst,color,bm,upgreek,booktabs,graphicx,multirow,warpcol}

\usepackage{times}
\usepackage{soul}
\usepackage{url}
\usepackage[hidelinks]{hyperref}
\usepackage[utf8]{inputenc}
\usepackage{amsmath}
\usepackage{amsthm}
\usepackage{booktabs}
\usepackage{algorithm}
\usepackage{algorithmic}
\usepackage[switch]{lineno}
\usepackage{bbm}
\usepackage{cite}

\usepackage{multirow}
\usepackage{amssymb}
\usepackage{booktabs}
\usepackage{color}
\usepackage{xcolor}
\usepackage{colortbl}
\usepackage{makecell}

\usepackage{amssymb}
\usepackage{extarrows}

\usepackage{epstopdf}

\newcommand{\red}[1]{{\color{black}#1}}
\newcommand{\blue}[1]{{\color{black}#1}}

\def\footnoterule{\kern 1mm \hrule width 10cm \kern 2mm}

\allowdisplaybreaks
\catcode`@=11
\def\title#1{\vspace{3mm}\begin{flushleft}\vglue-.1cm\Large\bf\boldmath\protect\baselineskip=18pt plus.2pt minus.1pt #1
\end{flushleft}\vspace{1mm} }
\def\author#1{\begin{flushleft}\normalsize #1\end{flushleft}\vspace*{-4pt} \vspace{3mm}}
\def\address#1#2{\begin{flushleft}\vglue-.35cm${}^{#1}$\small\it #2\vglue-.35cm\end{flushleft}\vspace{-2mm}\par}
\def\jz#1#2{{$^{\footnotesize\textcircled{\tiny #1}}$\let\thefootnote\relax\footnotetext{\!\!$^{\footnotesize\textcircled{\tiny #1}}$#2}}}
\catcode`@=11
\def\section{\@startsection{section}{1}{\z@}%
 {-3ex \@plus -.3ex \@minus -.2ex}%
 {2.2ex \@plus.2ex}%
{\normalfont\normalsize\protect\baselineskip=14.5pt plus.2pt minus.2pt\bfseries}}
\def\subsection{\@startsection{subsection}{2}{\z@}%
 {-3ex\@plus -.2ex \@minus -.2ex}%
 {2ex \@plus.2ex}%
{\normalfont\normalsize\protect\baselineskip=12.5pt plus.2pt minus.2pt\bfseries}}
\def\subsubsection{\@startsection{subsubsection}{3}{\z@}%
 {-2.2ex\@plus -.21ex \@minus -.2ex}%
 {1.4ex \@plus.2ex}
{\normalfont\normalsize\protect\baselineskip=12pt plus.2pt minus.2pt\sl}}

\begin{document}
\begin{CJK*}{GBK}{song}
\thispagestyle{empty}
\vspace*{-13mm}
\noindent {\small Journal of computer science and technology: Instruction for authors.
JOURNAL OF COMPUTER SCIENCE AND TECHNOLOGY}
\vspace*{2mm}

\title{FedEval-LLM: Federated Evaluation of Large Language Models on Downstream Tasks with Collective Wisdom}

\author{Yuanqin He$^{1}$, Yan Kang$^{1}$, Lixin Fan$^{1}$, and Qiang Yang$^{1,2}$}

\address{1}{AI Group, WeBank Co., Ltd, China}
\address{2}{Hong Kong University of Science and Technology, China}

\vspace{2mm}

\noindent E-mail: \{yuanqinhe, yangkang\}@webank.com;\\[-1mm]

\noindent {\small\bf Abstract} \quad  {\small Federated Learning (FL) has emerged as a promising solution for collaborative training of large language models (LLMs). However, the integration of LLMs into FL introduces new challenges, particularly concerning the evaluation of LLMs. Traditional evaluation methods that rely on labeled test sets and similarity-based metrics cover only a subset of the acceptable answers, thereby failing to accurately reflect the performance of LLMs on generative tasks. Meanwhile, although automatic evaluation methods that leverage advanced LLMs present potential, they face critical risks of data leakage due to the need to transmit data to external servers and suboptimal performance on downstream tasks due to the lack of domain knowledge. To address these issues, we propose a Federated Evaluation framework of Large Language Models, named FedEval-LLM, that provides reliable performance measurements of LLMs on downstream tasks without the reliance on labeled test sets and external tools, thus ensuring strong privacy-preserving capability. FedEval-LLM leverages a consortium of personalized LLMs from participants as referees to provide domain knowledge and collective evaluation capability, thus aligning to the respective downstream tasks and mitigating uncertainties and biases associated with a single referee. Experimental results demonstrate a significant improvement in the evaluation capability of personalized evaluation models on downstream tasks. When applied to FL, these evaluation models exhibit strong agreement with human preference and RougeL-score on meticulously curated test sets. FedEval-LLM effectively overcomes the limitations of traditional metrics and the reliance on external services, making it a promising framework for the evaluation of LLMs within collaborative training scenarios.}



\vspace*{4mm}

\end{CJK*}
\baselineskip=18pt plus.2pt minus.2pt
\parskip=0pt plus.2pt minus0.2pt
\begin{multicols}{2}

\section{Introduction}

Large Language Models (LLMs) have demonstrated remarkable performance across various general natural language processing (NLP) tasks. However, they often exhibit suboptimal performance when applied to downstream tasks of specific domains. This sparks the development of specialized LLMs that are trained on proprietary domain-specific data, including BloombergGPT~\cite{wu2023bloomberggpta} in the financial domain and HuaTuo~\cite{wang2023huatuo} in the biomedical field. To utilize these proprietary data while protecting data privacy, Federated Learning (FL)~\cite{mcmahan2017communicationefficient,yang2019federated} has emerged as a promising solution for the collaborative training of LLMs owing to its advanced privacy-efficiency trade-off~\cite{fan2023fatellm}. Since LLMs are primarily targeted at generative tasks, it is crucial to ensure a reliable evaluation of LLM's performance on downstream generative tasks. This evaluation is vital for overseeing the performance of the global model and assessing participants' contributions in various federated learning settings.

While numerous methods have been proposed for evaluating LLMs~\cite{liang2022holistic}, widely recognized evaluation methods for generative tasks remain limited~\cite{zhao2023surveyb}. Traditional methods typically leverage labeled test sets to evaluate generative tasks~\cite {chang2023survey,huang2023ceval}. However, these test sets, typically composed of one-to-one question-answer (QA) pairs, capture only a fraction of acceptable answers, thereby incapable of providing a reliable evaluation of LLM's performance~\cite{fu2023gptscore,liu2023geval}. This limitation is empirically demonstrated by our experiment presented in Table \ref{table:overview}, which shows that the performance of various LLMs measured by RougeL-score~\cite{lin2004rouge} reveals a significant dependence on the test datasets and does not align well with human preference. Moreover, constructing a labeled test set that covers the entire task scope of all participants within the FL framework raises risks of leaking private data.

Automatic evaluation methods employing advanced LLM, such as GPT4~\cite{openai2023gpt4}, exhibit promising agreement with human preference~\cite{fu2023gptscore,liu2023geval}, eliminating the need for labeled test sets. However, the deployment of such advanced LLMs within the FL framework is precluded as the need to transmit data to external servers poses serious risks to \textit{data privacy}. Furthermore, these general-purpose LLMs may exhibit suboptimal performance on downstream tasks due to the lack of \textit{domain knowledge}~\cite{wang2023huatuo,cui2023chatlaw}. For instance, tasks such as summarization or question-answering in specific domains may necessitate specific requirements on the granularity level of responses that are unattainable through in-context learning. 

As a result, the utilization of LLMs in FL necessitates the development of evaluation methods that can be deployed locally and are aligned with the specific requirements of downstream tasks~\cite{zheng2023judging,wang2023pandalm}. In this study, we term LLMs that can be deployed locally as "medium-sized LLMs," typically ranging in size from 3B to 30B, in contrast to their large counterparts, such as GPT4.

\tabcolsep 8pt
\renewcommand\arraystretch{1.3}
\begin{center}
{\footnotesize{\bf Table 1.} LLM performance measured by human judgment and RougeL score on 1K Alpaca instruction data~\cite{alpaca} and Dolly instruction data~\cite{DatabricksBlog2023DollyV2}}\\
\vspace{2mm}
\footnotesize{
\begin{tabular*}{\linewidth}{l||c|c|c}\hline\hline
		 \multirow{2}{*}{Model} & Rank*  & \multicolumn{2}{c}{RougeL} \\
          & By Human & On Alpaca & On Dolly \\
         \hline
         \hline
        Vicuna-13B~\cite{vicuna2023} & 1 & 37.8 &  26.5 \\ 
		\hline
		Koala-13B~\cite{koala_blogpost_2023}  & 2  & 33.7 &  23.9 \\ 
		\hline
        Pythia-12B~\cite{openpythia}  & 3  & 31.4  &  19.6 \\ 
		\hline
        ChatGLM-6B~\cite{zeng2022glm}  & 4 & 24.2 & 22.4\\ 
		\hline
        StableLM-7B~\cite{StableLMAlphaV2Models} & 5 & 25.6 & 16.9 \\ 
		\hline
        Dolly-12B~\cite{DatabricksBlog2023DollyV2}  &  6 & 31.9 & 51.9 \\ 
		\hline
        LLaMA-13B~\cite{touvron2023llamaa}  & 7 &  23.0 & 20.4 \\ 
     \hline\hline\hline
	\end{tabular*}
\\\vspace{1mm}\parbox{8.3cm}{Rank* is obtained from ~\cite{zheng2023judging}. A higher RougeL score means a better match with the reference answer. Pythia-12B stands for OpenAssistant-Pythia-12B and StableLM-7B stands for StableLM-Tuned-Alpha-7B.}
}
\label{table:overview}
\end{center}

\vspace{1mm}

\begin{figure*}[ht!]
	\centering
\includegraphics[width=0.95\textwidth] {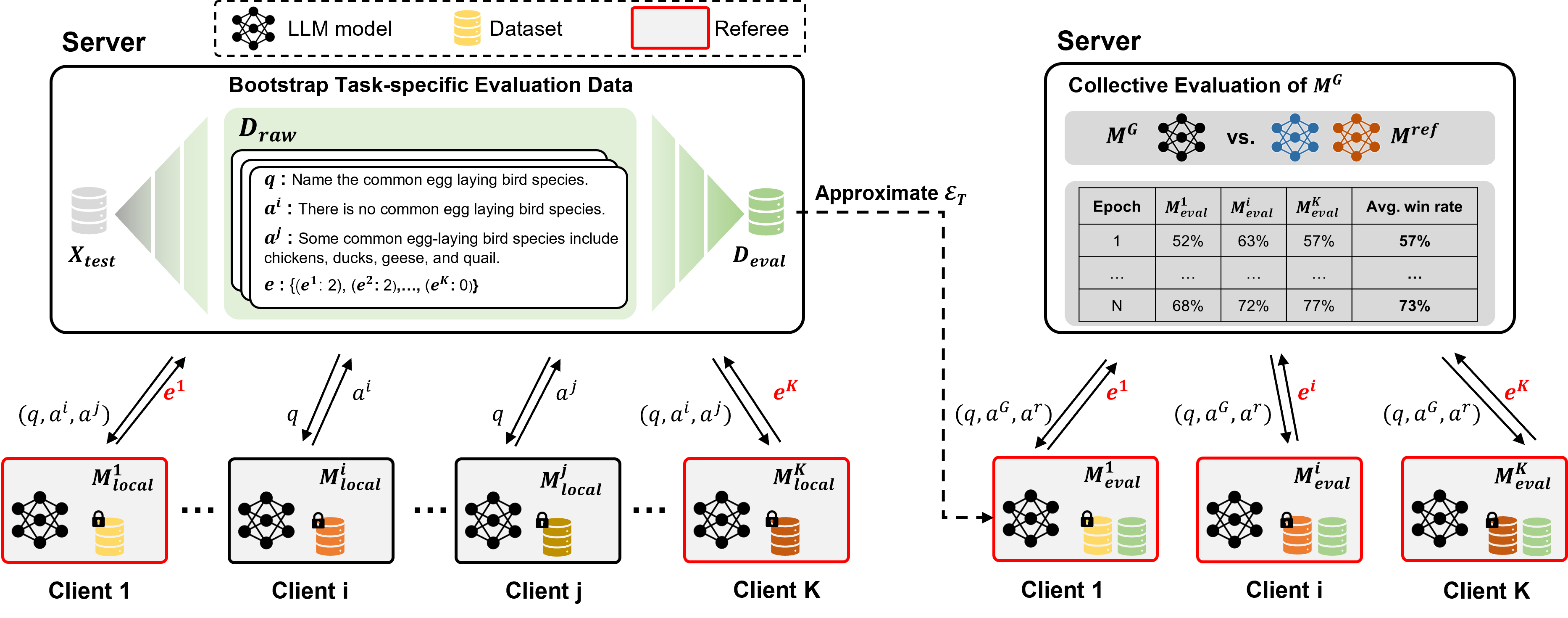}
	\caption{Overview of the proposed FedEval-LLM framework. It presents two key steps: (1) Training of personalized evaluation models (left) and (2) Collective evaluation using these evaluation models (right). \textbf{Left}: With well-trained local models, $M^i_{local}$, participating clients build a task-specific evaluation dataset $D_{eval}$ based on a question-only dataset $X_{test}$ utilizing a bootstrapping strategy. The obtained evaluation dataset serves as an approximation of the task-specific evaluation criteria $\mathcal{E}_T$, and is used to train a personalized evaluation model for each client, $M^i_{eval}$. Here, $(q, a)$ and $e$ represent the question-answer pair and the corresponding evaluation. \textbf{Right}: In the training phase, a group of clients acts collectively as referees, providing a reliable evaluation of the global model, $M^G$, as the win rate to the reference models, $M^{ref}$. A detailed description of the framework is given in Section 3.} 
	\label{fig:overview}
\end{figure*}

To tackle this problem, we propose FedEval-LLM, a \underline{Fed}erated \underline{Eval}uation framework for \underline{LLM} (depicted in Fig.~\ref{fig:overview}), designed to provide reliable performance assessment of LLMs on downstream tasks. Specifically, we utilize fine-tuned evaluation models from participants serving as referees, thus overcoming the requirement for labeled test data and mitigating privacy concerns associated with external tools. These personalized evaluation models are capable of following the specific evaluation criterion of the downstream task by leveraging the domain knowledge of participants through their distinct local data and bootstrapped high-quality evaluation data. Moreover, superior evaluation performance on downstream tasks is achieved by using a group of evaluation models as referees, encompassing distinct local perspectives of clients. We demonstrate that utilizing a group of LLMs as referees is essential for both selecting high-evaluation data and providing reliable evaluation. Consequently, by obviating the need for external services and labeled test sets we assert that FedEval-LLM provides strong privacy-preserving capability compared to other methods. Our contributions can be summarized as follows:
\begin{itemize}
    \item We make a first attempt in FL to develop the task-specific evaluation capability by leveraging participating LLM's domain knowledge and collective wisdom on downstream tasks. Our investigation delves into the various factors that contribute to the evaluation capability on downstream tasks, highlighting the pivotal role of domain knowledge in adhering to the relevant evaluation criteria. The experimental results confirm the capability of the proposed FedEval-LLM framework, demonstrating its competitive performance, which is comparable to the most advanced local evaluation models. 
    \item We apply it to a federated learning scenario involving eight clients on two distinct downstream tasks. The evaluation results exhibit strong agreement with RougeL-score on a carefully curated labeled test set and human preference, demonstrating the effectiveness of the proposed FedEval-LLM framework.
\end{itemize}

\section{Related Works}

\subsection{Fine-tuning of Large Language Models}
Despite the ever-increasing general capability of LLMs~\cite{openai2023gpt4}, their performance on downstream tasks is often limited due to a lack of domain knowledge~\cite{wang2023huatuo,cui2023chatlaw}. While this challenge can be partially addressed through in-context learning using exemplars or augmented capabilities~\cite{mialon2023augmenteda}, fine-tuning LLMs on the target dataset is considered the more effective solution. Research has demonstrated that fine-tuning medium-sized LLMs can surpass more advanced LLMs in various tasks, including recommendation~\cite{bao2023tallrec} and information extraction~\cite{dunn2022structured}.

Parameter-efficient fine-tuning (PEFT) techniques have been widely adopted to fine-tune LLMs to balance the training cost and model performance~\cite{ding2022delta}. These techniques typically train a small amount of parameters while leaving the rest frozen, which significantly reduces the computational and storage requirements, thereby making fine-tuning LLMs more affordable. Prominent PEFT methods includes LoRA~\cite{hu2021lora}, P-Tuning~\cite{liu2021ptuninga}, and Prompt Tuning~\cite{lester2021powera}. In this work, we opt for LoRA, given its proven effectiveness and compatibility with various model architectures.

\subsection{Evaluation of LLMs}

The diverse capability of LLMs and their generative nature make the evaluation of LLMs extremely challenging~\cite{chang2023survey}. Traditional methods rely on carefully curated test sets to evaluate the model performance on specific task~\cite{liang2022holistic,zhong2023agieval}, ranging from simple sentiment classification to more diverse Multilingual Multi-Genre Language Understanding (MMLU) benchmarks~\cite{hendryckstest2021}. Quantitative measures for generative tasks predominantly hinge on similarity-based metrics, including text-wise~\cite{lin2004rouge} and semantic-wise~\cite{zhang2020bertscorea} approaches. However, recent research indicates that these methods exhibit subpar performance in open-ended generative tasks~\cite{wang2023chatgpt}, and become increasingly challenging to implement due to the necessity for golden labels.

Recently, automatic evaluation using advanced LLMs (e.g., GPT4~\cite{openai2023gpt4}) as the referee has been explored and adopted in several works~\cite{fu2023gptscore}. Zheng \textit{el al.} \cite{zheng2023judging} demonstrated that advanced LLMs exhibit strong agreement with human preference on general tasks. However, there have been limited works that apply this approach to downstream tasks~\cite{wang2023chatgpt,chang2023survey}.  In addition, medium-sized LLMs, which are locally deployable, such as LLaMA~\cite{touvron2023llamaa},  exhibit limited evaluation performance~\cite{wang2023pandalm,zheng2023judging}. It remains unexplored how to effectively impart the evaluation capability to these models.

\subsection{Federated Learning with LLM}
Federated learning (FL) is a privacy-preserving paradigm that enables multiple participants to collaboratively train machine learning models while protecting private data~\cite{mcmahan2017communicationefficient,yang2019federated}. It presents greater flexibility in balancing efficiency and privacy, making it particularly well-suited for tackling challenges encountered in LLM development and deployment, including the scarcity of high-quality data, data privacy concerns, and constraints imposed by limited computational resources~\cite{kang2023grounding}. 

FL has been widely adopted in Natural Language Processing (NLP) tasks~\cite{zhang2023whena}. 
Pre-trained Language models (PLM) have been introduced as the initial clients' models to enhance the convergence and mitigate data-heterogeneity problem~\cite{guo2022promptfla,zhao2023fedprompt}. In a recent study~\cite{zhang2023building}, distributed instruction tuning data was leveraged to train a global model using FL, relying exclusively on the GPT4-based evaluation method without resorting to similarity-based metrics. This further underscores the limitations of such metrics and highlights the need for more suitable evaluation methods.


In this section, we first formulate the evaluation problem, and then we elaborate on our proposed framework.

\subsection{Problem Formulation}
We consider the \blue{horizontal} federated learning setting, in which $K$ \blue{clients}, each owning a \blue{local model}, $M^i$ parameterized by $\theta^i$, intend to collaboratively train a global LLM, $M^G$ parameterized by $\theta^G$, on the target generative task, $T$ using their private data, $\{(X^i, Y^i)\}_1^K$. 

\noindent{\bf{Objective of Federated Learning.}} The objective of the conventional FL (i.e., FedAvg~\cite{mcmahan2017communicationefficient}) is to train a global model by minimizing the following global loss:
\begin{equation}
    \min_{\theta^G} ~ \frac{1}{K} \sum_{i=1}^K \mathcal{L}^i_T (Y^i,M^i(X^i)).
\end{equation}
where $\mathcal{L}^i_{T}$ is the loss function of the $i$-th client on the target task, $T$, and $\theta^i$ is initialized with $\theta^G$. 

FedAvg iteratively aggregates clients' local updates trained on local data to update the global model. In this way, the local update serves as a proxy for harnessing each client's local knowledge.

\noindent{\bf{Objective of Federated Evaluation.}} Regarding the evaluation, as elaborated in previous sections, the goal is to construct a locally deployable evaluation model, $M^G_{eval}$ parameterized by $\theta^G_{eval}$, while preserving local data privacy. This evaluation model should be capable of evaluating the target task reliably. In particular, given a question, $x$, and the corresponding answer, $y$, generated by some LLMs, $M_{eval}$ evaluates how good the answer $y$ is given the question $x$
\begin{equation}
    e = M^G_{eval}(y|x).
\end{equation}
where $e$ is the evaluation output and its format depends on the chosen evaluation method~\cite{zheng2023judging}. In a single answer grading method, $e$ is a grading score to a single answer assigned by $M^G_{eval}$, while in the pairwise comparison method where $y$ comprises two answers, $e$ represents the index of the preferred answer determined by $M^G_{eval}$. 

Downstream generative tasks often come with different requirements in terms of aspects, lengths, and formats of the answer, depending on the application scenarios. For example, instruction tuning tasks focus on the \red{helpfulness and relevance} of the answer, while summary tasks pay more attention to the content coherence to the original text and the length of the answer. As a result, the evaluation of respective tasks should take these requirements into account and reflect the corresponding task preference. For example, answers with more details will be given higher scores when evaluating the instruction tuning task but not preferred by the summary task. We refer to this task preference in the context of evaluation as \textit{task-specific evaluation criteria} $\mathcal{E}_T$, where the suffix $T$ highlights the task dependency. Consequently, \blue{the objective of the federated evaluation is to train a global evaluation model}, $M^G_{eval}$, that is aligned with $\mathcal{E}_T$:
\begin{equation}
    \min_{\theta^G_{eval}} ~ \frac{1}{K} \sum_{i=1}^K \mathcal{L}^i_{E} (\mathcal{E}_T, M^i) 
\end{equation}
where $\mathcal{L}^i_{E}$ is the loss function of the $i$-th client on the evaluation task, and $\theta^i$ is initialized with $\theta^G_{eval}$.  

\subsection{Motivation}
\red{It's important to note that the task-specific evaluation criteria of generative tasks, $\mathcal{E}_T$, is hard to quantify and not explicitly available}, thus making its alignment challenging. This is similar to the task of aligning LLMs with human preference, where human-labeled comparison data is used by the Reinforcement Learning from Human Feedback (RLHF) method~\cite{ouyang2022trainingb} to approximate the human preference. 

To align $M^G_{eval}$ with the task-specific evaluation criteria, $\mathcal{E}_T$, we propose to build a task-specific evaluation dataset, $D_{eval} = ((X_{T}, Y_{T}), E_{T})$ to approximate $\mathcal{E}_T$, where ($X_{T}$, $Y_{T}$) are question-answer pairs of target task and $E_{T}$ is the ground-truth evaluation output of $Y_{T}$ given $X_{T}$. The suffix $T$ emphasizes that they are all \textit{task-specific}. In this way, we transformed the implicit evaluation criteria (task preference) into explicit input-output pairs that can be used in supervised fine-tuning. Therefore, the objective can be reformulated as 
\begin{equation}
    \min_{\theta^G_{eval}} ~ \frac{1}{K} \sum_{i=1}^K \mathcal{L}^i_{E} (E_{T}, M^i_{eval}(Y_{T}|X_{T})).
    \label{eq:eval_v2}
\end{equation}

To ensure that the evaluation dataset adheres to $\mathcal{E}_T$, a bootstrapping strategy that leverages the multi-task processing capability of LLMs is employed. \red{Specifically, we utilize local model of each client $i$, $M^i_{local}$ parameterized by $\theta^i_{local}$},  to harness the knowledge of the target task through training on local data, $(X^i, Y^i)$,
\begin{equation}
    \theta^i_{local} \leftarrow \theta^i - \nabla_{\theta^i} \mathcal{L}^i_T (Y^i, M^i(X^i)).
    \label{eq:local_llm}
\end{equation}
Here, the learning rate is omitted for simplicity. These local models, $M^i_{local}$, capable of following the requirement of the target task, are employed for both the downstream task, generating $Y_{T}$ given $X_T$, and the evaluation task as referees, providing evaluation $E_{T}$ on ($X_T, Y_T$). This dual role ensures that the resulting $D_{eval} =((X_{T}, Y_{T}), E_{T})$ closely aligns with the task preference.

While the $D_{eval}$ facilitates the training of a global evaluation model, as depicted in Eq. \ref{eq:eval_v2}, it faces two limitations. First, the $D_{eval}$, obtained by the bootstrapping method, predominantly reflects the shared preference of participating clients, neglecting the distinct local perspective of individual clients. This may lead to an overly narrow evaluation scope, potentially resulting in degenerated evaluation capability. Second, a single evaluation model may suffer from limited evaluation capability due to the restricted number of evaluation data and the relatively modest model size.

To address these limitations, we propose to leverage the local knowledge of clients by training personalized evaluation models, $M^i_{eval}$, for each client based on their respective local models, $M^i_{local}$,
\begin{equation}
    \min_{\theta^i_{eval}} ~ \mathcal{L}^i_{E} (E_{T}, M^i_{eval}(Y_{T}|X_{T})),i=1,\dots,K,
\end{equation}
where $\theta^i_{eval} \leftarrow \theta^i_{local} - \nabla_{\theta^i_{local}} \mathcal{L}^i_E (E_{T}, M^i_{local}(Y_{T}|X_{T}))$. 

This approach ensures that each personalized evaluation model not only acquired the common task-specific evaluation criteria from $D_{eval}$ but also incorporated the distinct local perspective through the training on their local data.

For the evaluation, we use a group of $L$ personalized evaluation models as referees, $M_{eval} = \{M^i_{eval}\}_1^L$, to overcome the constraints of a single model and broaden the evaluation scope. Given a question-answer pair, we adopt a majority voting strategy to aggregate the evaluation results from multiple referees, yielding the evaluation as follows
\begin{equation}
        e_{final} = \text{majority\_voting}(\{M^i_{eval}(y|x)\}_1^L).
\end{equation}
This approach has two notable advantages: 1) The utilization of multiple referees can provide \textit{enhanced reliability} by reducing the variance arising from the limited evaluation capability of a single model. 2) Individual evaluation models contribute unique evaluation perspectives compared to the global model trained solely on $D_{eval}$, thereby offering \textit{diversified evaluation scope}. 

Consequently, by leveraging the collective wisdom of participating clients, we transform local knowledge into task-specific evaluation models. Utilizing these evaluation models as a collective group provides reliable evaluation capability on the target task under the FL framework. Importantly, this obviates the requirement for labeled test sets and external services, thereby providing stronger privacy-preserving capability.

Next, we explain these two steps in detail with a focus on the training of personalized evaluation models.

\subsection{Training Personalized Evaluation Model}
The key to obtaining personalized evaluation models on downstream tasks lies in the effective utilization of domain knowledge dispersed among participants and its transformation into an evaluation capability that is tailored to the pertinent task. The main idea is to construct a high-quality evaluation dataset, $D_{eval}$, that aligns with the evaluation criteria of the downstream task. 

To protect the privacy of local data, we introduce a separate test set, $X_{test}$, comprising only questions. This test set can be constructed collaboratively by all participants or derived from publicly available data within a similar domain.

The process is delineated into the subsequent stages and illustrated in the Fig. 1.

\begin{enumerate}
    \item Training Local LLM. To harness the distributed domain knowledge in a privacy-preserving manner, each participant initiates by training a local LLM, $M^i_{local}$ using their local data, $(X^i, Y^i)$, $i=1,\dots, K$.
    \item Bootstrapping Evaluation Data. In this step, the local LLM is employed for both the downstream task and the evaluation task. To acquire evaluation data, participants engage in pairwise competition, employing their own local LLM on randomly sampled questions from the test set, $X_{test}$. This can maximally exploit the local knowledge of all participants. Specifically, for a sampled question $q$ and two participants, client $i$ and client $j$, we generate a question-answer (QA)-pair, $(q, (a^i, a^j))$, where $a^i = M^i_{local}(q)$ and $a^j = M^j_{local}(q)$. The QA-pair is subsequently evaluated by $N$ other participants utilizing their local LLM, resulting in a complete evaluation sample, $(q, (a^i, a^j), (e^1,\dots,e^N))$, where $e^n = M^{n}_{local}(q, (a^i, a^j))$ and $n \neq i,j$. This generation step may be repeated multiple times to construct an evaluation dataset, $D_{raw}$.
    \item Selecting High-Quality Evaluation Data. Given that raw evaluation samples may be noisy due to the limited evaluation capability and knowledge of individual local LLM, we apply three levels of consensus to select high-quality data, i.e., \textit{order-consistency}, \textit{output-consistency}, and \textit{judgment-consistency}. Further elaboration on these criteria is provided later in this section. Then we obtain a set of high-quality evaluation data, denoted as $D_{eval} = \text{filter}(D_{raw})$. 
    \item Each participant proceeds to train a personalized evaluation model, $M^i_{eval}$, based on their local LLM, $M^i_{local}$, and the selected evaluation data, $D_{eval}$. Consequently, the obtained evaluation models encapsulate both the distinct local knowledge and a "common" evaluation capability on the downstream task.
\end{enumerate}

We explain some details mentioned in the above steps.

\noindent{\bf{Evaluation Protocol.}}
Considering the limited evaluation capability of medium-sized LLMs, we use the more simple evaluation paradigm, i.e., \textit{pairwise comparison}\cite{wang2023pandalm,zheng2023judging}. Specifically, given two answers, $a_1$ and $a_2$, for the question $q$, the referee LLM outputs its evaluation $e$, which indicates which answer it thinks is better, "1" means a preference for the first answer, "2" for the second answer, and "0" indicates no preference between them. 

\noindent{\bf{Selection Criteria.}} Three criteria are employed to improve the quality of the selected evaluation data: 1) \textit{order-consistency}, 2) \textit{output-consistency}, and 3) \textit{judgment-consistency}. 

First, we utilize order consistency to address order preferences, as reported in \cite{zheng2023judging}. Specifically, we present QA-pairs to LLM referees in both possible orders, namely $e_{ord}=M(q, (a^i, a^j))$ and $e_{inv}=M(q, (a^j, a^i))$. Only those with consistent judgment are considered as \textit{valid} and will be utilized. It's worth noting that, unlike previous work\cite{zheng2023judging}, where inconsistent evaluations were treated as a 'tie', in our approach, they are discarded, aiming to provide more reliable results by minimizing noise.

Then, given the valid evaluations of multiple referees, output and judgment consistencies are utilized to further enhance the data quality. Output consistency is defined as the proportion of valid evaluations to the total number of evaluations, while judgment consistency relates to the fraction of valid evaluations with matching judgments. These two criteria aim to eliminate low-confidence samples. Although this leads to a reduced quantity of usable samples, we can compensate for this by conducting additional rounds of generation steps to obtain the necessary quantity of samples. 

\noindent{\bf{Adaptive Training Strategy.}} To expedite the sample collection process, we introduced an adaptive strategy that progressively refines the selection criteria based on the number of acquired evaluation samples. Specifically, we gradually apply these three criterion. Initially, we employ the order-consistency criterion, swiftly accumulating valid evaluation samples, which are used to train temporary personalized evaluation models. This temporary evaluation model, capable of producing a higher volume of valid samples, is then deployed for the subsequent rounds of data generation. Finally, the entire evaluation data is filtered by all three criteria, further enhancing the quality of the selected samples. 

\subsection{Collective Evaluation}
\label{subsec:collective}

These personalized evaluation models can be integrated into the FL framework to assist in monitoring training procedures and optimizing training objects. They are used collectively to mitigate bias and uncertainty encountered by a single LLM acting as a referee. To track training progress, we measure the relative performance of the global model compared to predefined reference models, including base models and local models of participating clients. This same approach can also be applied to determine the individual contribution of local models.

For the collective evaluation, only order-consistent results are considered, i.e., valid evaluations and majority voting are employed to determine the outcome of each question (win or lose). To account for the confidence of each question, we weight it based on the ratio of the number of win (lose) evaluations to the number of referees. This ensures that evaluation with more valid and consistent judgment receives higher importance. Subsequently, the scores are summed over the winning and losing questions, respectively. The final result is calculated as the fraction of the sum of the weighted win-score (or lose-score) divided by the total weighted score.

\section{Experiments}
\label{sec:experiment}
\subsection{Experimental Settings}
\noindent{\textbf{Federated Learning Setting.}} 
We simulate a federated learning scenario involving eight clients using LLaMA-7B~\cite{touvron2023llamaa} as the base model. We follow the native horizontal federated learning protocol, which leverages the proportion of local data as aggregation weights~\cite{mcmahan2017communicationefficient}. We conduct 10 communication rounds, with one local epoch per each communication round. 

For both the training of local models and personalized evaluation models, we adopt the Low-Rank Adaption method (LoRA)~\cite{hu2021lora} and construct the training pipeline based on the Alpaca-lora repository\jz{1}{https://github.com/tloen/alpaca-lora}. We employ the same hyperparameters for both local and evaluation model training, as listed in Table 2.

We employ two distinct downstream tasks: an instruction tuning task and a summary task. We select eight publicly available datasets covering diverse domains and languages for the instruction tuning task. Each client is assigned 5K samples randomly drawn from one of these datasets. The chosen datasets include Dolly~\cite{DatabricksBlog2023DollyV2}\jz{2}{https://github.com/databrickslabs/dolly}, Alpaca~\cite{alpaca}, Self-Instruct~\cite{wang2022selfinstruct}, HC3-English~\cite{guo2023howa}\jz{3}{https://huggingface.co/datasets/Hello-SimpleAI/HC3}, HC3-Chinese~\cite{guo2023howa}\jz{4}{https://huggingface.co/datasets/Hello-SimpleAI/HC3-Chinese}, BELLE-Math~\cite{belle2023exploring}\jz{5}{https://huggingface.co/datasets/BelleGroup/school$\_$math$\_$0.25M}, BELLE-1M-Chinese~\cite{belle2023exploring}\jz{6}{https://huggingface.co/datasets/BelleGroup/train$\_$1M$\_$CN} and 
Guanaco\jz{7}{https://huggingface.co/datasets/JosephusCheung/GuanacoDataset}. For the summary task, we utilize the TL;DR dataset~\cite{volske2017tl} and assign each client with 10K samples.

\noindent{\textbf{Training Personalized Model.}} To select high-quality evaluation data, we let participating LLMs compete with each other until a specified quantity of evaluation samples is attained. Each competition involves a set of 100 questions and is evaluated by a group of 5 referees. The prompt templates for training local LLMs and personalized evaluation models can be found in Table 3. We adopt a two-stage training strategy for both datasets. In the first generation round, evaluation data is collected with relatively loose criteria to train temporary evaluation models. In the subsequent four generation rounds, the evaluation data obtained by temporary evaluation models are filtered with tightened selection criteria to enhance the quality of the evaluation samples.

\setcounter{table}{2}
\begin{table*}[ht]
\small
\caption{Prompt template used for local model and personalized evaluation model training.}
\centering
    \begin{tabular}{{p{3cm}p{12.5cm}}}
        \hline
        ~ & \textbf{Prompt template} \\
        \hline
        \textbf{Local model training} & \makecell[l]{Below is an instruction that describes a task, paired with an input that provides further context. \\ Write a response that appropriately completes the request. \\\\
        \textbf{Instruction}: $\{$instruction$\}$ \\
        \textbf{Input}: $\{$input$\}$ \\
        \textbf{Response}: } \\
        \hline
        \textbf{Personalized evaluation model training (Instruction tuning)} & \makecell[l]{Below are two responses for a given task. Evaluate the responses and decide which response is \\better in terms of helpfulness, relevance, accuracy, and level of detail. Output a single line \\containing only one value indicating the number of responses you think is better. Use '1' to \\represent that response 1 is better, '2' to represent that response 2 is better, or 'tie' if you think \\the results are similar or inconclusive. Avoiding any potential bias and ensuring that the order \\in which the responses are presented does not affect your judgment. \\\\
        \textbf{Instruction}: $\{$instruction$\}$ \\
        \textbf{Input}: $\{$input$\}$ \\
        \textbf{Response 1}: $\{$answer 1$\}$ \\
        \textbf{Response 2}: $\{$answer 2$\}$ \\
        \textbf{Evaluation}: } \\
        \hline
        \textbf{Personalized evaluation model training (TL;DR summary)} & \makecell[l]{Below are two summaries of a given article. Evaluate the summaries and decide which summary \\is better in terms of coherence, relevance, consistency, and fluency. Output a single line containing \\only one value indicating the number of the summary you think is better. Use '1' to represent that \\response 1 is better, '2' to represent that response 2 is better, or 'tie' if you think the results are \\similar or inconclusive. Avoiding any potential bias and ensuring that the order in which the \\responses are presented does not affect your judgment. \\\\
        \textbf{Instruction}: $\{$instruction$\}$ \\
        \textbf{Input}: $\{$input$\}$ \\
        \textbf{Response 1}: $\{$answer 1$\}$ \\
        \textbf{Response 2}: $\{$answer 2$\}$ \\
        \textbf{Evaluation}: } \\
        \hline
    \end{tabular}
\label{table:prompt_template}
\end{table*}

\tabcolsep 10pt
\renewcommand\arraystretch{1.3}
\begin{center}
{\footnotesize{\bf Table 2.} Hyperparameters used for fine-tuning.}\\
\vspace{2mm}
\footnotesize{
\begin{tabular*}{\linewidth}{c|l|c}\hline\hline\hline
		& Parameter & Value \\ 
	    \hline
        \hline
        \multirow{3}{*}{Federated Learning} & Number of client & 8  \\
        & Communication round & 10  \\
        & Local epoch & 1  \\
        \hline
        \multirow{6}{*}{Fine-tuning} & Learning rate & 1e-4  \\
        & Batch size & 16   \\
        & Max Seq. Len. & 1280 \\
        & LoRA $\alpha$ & 16 \\
        & LoRA r & 16 \\ 
		& LoRA dropout  & 0.05  \\
        & Max. length of new tokens  & 512  \\
        \hline
        \multirow{2}{*}{Inference} & Temperature (answer) & 0.7  \\
        & Temperature (referee) & 0 (greedy)   \\
        \hline\hline\hline
	\end{tabular*}
}
\label{table:hyperparam}
\end{center}

\noindent{\textbf{Evaluation Metrics.}} To assess the evaluation capability of LLM, we measure their agreement on human-annotated test sets. Specifically, we utilize the test set derived from PandaLM~\cite{wang2023pandalm} for the instruction tuning task and randomly sampled 1K comparison data from the test set introduced by Stiennon \textit{et al.}~\cite{stiennon2022learning} for the TL;DR summary. We employ two key metrics to evaluate the model's performance: the fraction of valid evaluations, denoted as $r_v$, and the accuracy of these valid evaluations, denoted as $Acc_v$. These metrics can be mathematically formulated as follows
\begin{equation}
    r_v = \frac{e_{valid}}{e_{total}}, ~~ 
    Acc_v = \frac{e_{correct}}{e_{valid}}, 
\end{equation}
where $e_{total}, e_{valid}$ and $e_{correct}$ represent the total evaluations, valid evaluations, and correct evaluations, respectively, as depicted in Fig. 2. 

$r_v$ and $Acc_v$ encompass distinct aspects of the evaluation capability: $r_v$ indicates the \textit{inherent evaluation capability}, which measures its instruction following capability on the evaluation task, while $Acc_v$ represents the LLM's \textit{task-related evaluation capability}, indicating the degree of alignment with the downstream task. We further define accuracy $Acc_t = r_v \cdot Acc_v$, which provides the overall measure of the model's evaluation capability on the downstream task. 

\setcounter{figure}{3}
\begin{center}
\includegraphics[width=6cm]{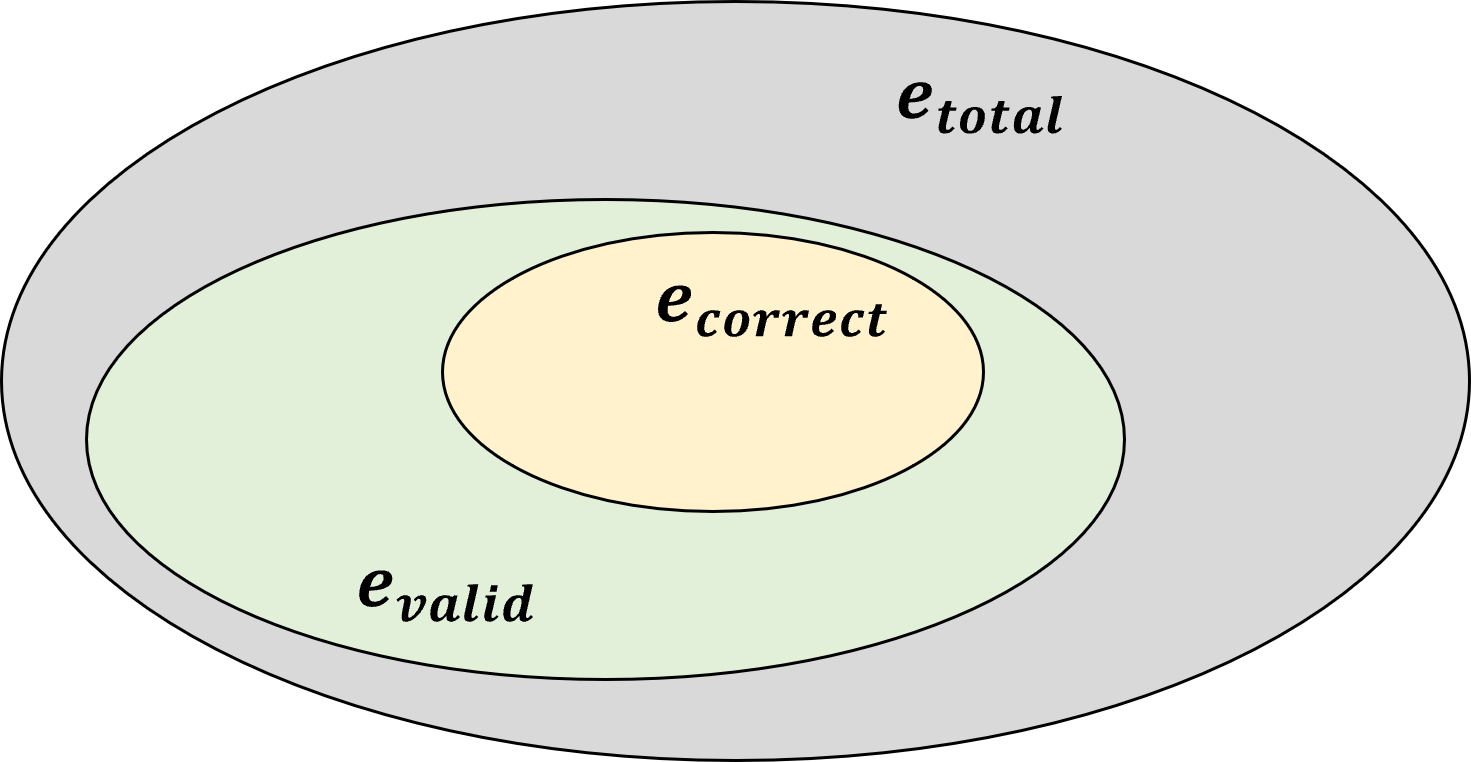}\\
\vspace{3mm}
\parbox[c]{8.3cm}{\footnotesize{Fig.2.~} Schematic illustration of evaluation space. The gray region, green region, and yellow region represent total evaluations, valid evaluations, and correct evaluations, respectively.}
\label{fig:eval_space}
\end{center}

\subsection{Experimental Results}

\subsubsection{Personalized Evaluation Model}

\noindent{\textbf{Enhanced Evaluation Capability of Individual Evaluation Models.}} 
We compare the evaluation capability of the local model and personalized evaluation model by measuring their agreement on human-annotated test sets. The results, obtained by averaging over 8 clients, are presented in Table \ref{table:eval_ft}. The original LLaMA-7B exhibits notably low values for both $r_v$ and $Acc_v$, suggesting limited instruction following capability on evaluation tasks and lack of knowledge on downstream tasks. Following fine-tuning on local data, the local LLM (Local LLM (avg)) gains a certain degree of evaluation capability, as evidenced by an increase in $Acc_v$. However, the low $r_v$ suggests the model's inherent evaluation capability remains quite constrained. With the FedEval-LLM method (FedEval-LLM (avg)), we observe a significant improvement of $r_v$, from $0.149$ to $0.628$ for the instruction tuning task, and from $0.016$ to $0.484$ for the TL;DR summary task. Along with a slight enhancement of $Acc_v$, FedEval-LLM achieves the best overall evaluation capability, demonstrating the efficacy of the proposed method and the high quality of the selected evaluation dataset. 

\setcounter{table}{3}
\begin{table*}[th!]
\small
\caption{Evaluation capability measured on human-annotated datasets. Local LLM (avg) denotes the average result of LLMs fine-tuned on local data. LLaMA-7B-Eval denotes the LLaMA-7B fine-tuned on evaluation data. }
\centering
\begin{tabular}{l|ccc|ccc}
        \hline
		Dataset & \multicolumn{3}{c|}{Instruction tuning} & \multicolumn{3}{c}{TL;DR} \\ 
        \hline
		Model & $r_v$ & $Acc_v$ & $Acc_t$ & $r_v$ & $Acc_v$ & $Acc_t$ \\ 
	    \hline
        \hline
        LLaMA-7B & 0.136 & 0.434 & 0.059 & 0.025 & 0.560 & 0.014  \\
        Local LLM (avg) & 0.149 & 0.696 & 0.104 & 0.016 &  0.615 & 0.010  \\
        LLaMA-7B-Eval & 0.530 & 0.707 & 0.374 & 0.387 & 0.605 & 0.234 \\
        PandaLM & 0.718 & \textbf{0.713} & 0.512 & 0.468 & 0.521 & 0.244 \\  
        \hline
		FedEval-LLM (avg) & 0.628 & 0.710 & 0.446 & 0.484 & \textbf{0.623} & 0.302 \\
        FedEval-LLM (3-referees) & 0.839 & 0.660 & 0.554 &
        0.786 & 0.594 & 0.467  \\
        FedEval-LLM (5-referees) & \textbf{0.887} & 0.650 & \textbf{0.577} &\textbf{0.853} & 0.580 & \textbf{0.495}  \\
        \hline\hline\hline
	\end{tabular}

\label{table:eval_ft}
\end{table*}

\noindent{\textbf{Optimal Performance Achieved through Collective Evaluation.}} We investigate the evaluation capability utilizing a group of LLMs as referees, following the evaluation rules described in Section 3.3. The results, shown in the last two rows of Table \ref{table:eval_ft}, are obtained by averaging the evaluation results conducted by randomly selected $N$ LLMs ($N = 3$ or $5$), each repeated 20 times. Compared to directly fine-tuning the base model (LLaMA-7B-Eval), it is evident that the metric $r_v$ exhibits a significant increase for both datasets, rising from $0.628$ to over $0.839$ for the instruction tuning task and from $0.484$ to $0.786$ for the TL;DR summary task when employing three referees (FedEval-LLM (3-referees)). We think this improvement is attributed to the distinct distribution of valid evaluations generated by personalized evaluation models, suggesting the preservation of diverse local data knowledge, which is beneficial for the collective evaluation. Otherwise, the outputs tend to align closely in judgment, resulting in a marginal increase of $r_v$ compared when using a single referee. Furthermore, a slight reduction of $Acc_v$ also supports our suggestion, which indicates discrepancies in judgments among referees. Overall, this underscores that collective evaluation, employing multiple LLMs, yielded the best evaluation performance, as evidenced by the highest $Acc_t$, which is $0.577$ for the instruction tuning task and $0.495$ for the TL;DR summary task when five referees are employed (FedEval-LLM (5-referees)). 

\noindent{\textbf{Superior Adaptability to Downstream Tasks.}} We proceed to compare the performance of FedEval-LLM to PandaLM~\cite{wang2023pandalm}, a specialized evaluation model trained on 300K instruction tuning evaluation data. This specialized model can be regarded as a potent local substitute for the advanced LLM on automatic evaluation. For the instruction tuning task, PandaLM manifests superior performance when utilized as a single referee, while FedEval-LLM demonstrates comparable results. For the TL;DR summary dataset, PandaLM continues to exhibit relatively high $r_v$, indicating its good inherent evaluation capability, which has been gained through training on evaluation data. However, the low $Acc_v$ suggests its evaluation criterion is not aligned with the downstream task, thus resulting in a suboptimal $Acc_t$. This misalignment primarily arises from the lack of knowledge related to the downstream task. In contrast, FedEval-LLM, which utilizes the domain knowledge from the target task, demonstrates enhancement in both $v_r$ and $Acc_v$. This confirms our assertion that domain knowledge is essential for the accurate evaluation of downstream tasks.

\subsubsection{Evaluation in Federated Learning}

We demonstrate the evaluation capability of the proposed framework on two aspects that are most important in FL, the performance of the global model during the training and performance differences among local models. 

\noindent{\textbf{Global Model Evaluation.}} We conducted a federated learning involving 8 clients. To construct a global test set, we selected 100 samples from each client's test dataset, that is 800 test samples in total. For FedEval-LLM, we randomly selected 3 referees and employed the evaluation results of PandaLM and RougeL-score for comparison, as presented in Fig. 3. 

\setcounter{figure}{3}
\begin{center}
\includegraphics[width=8.5cm]{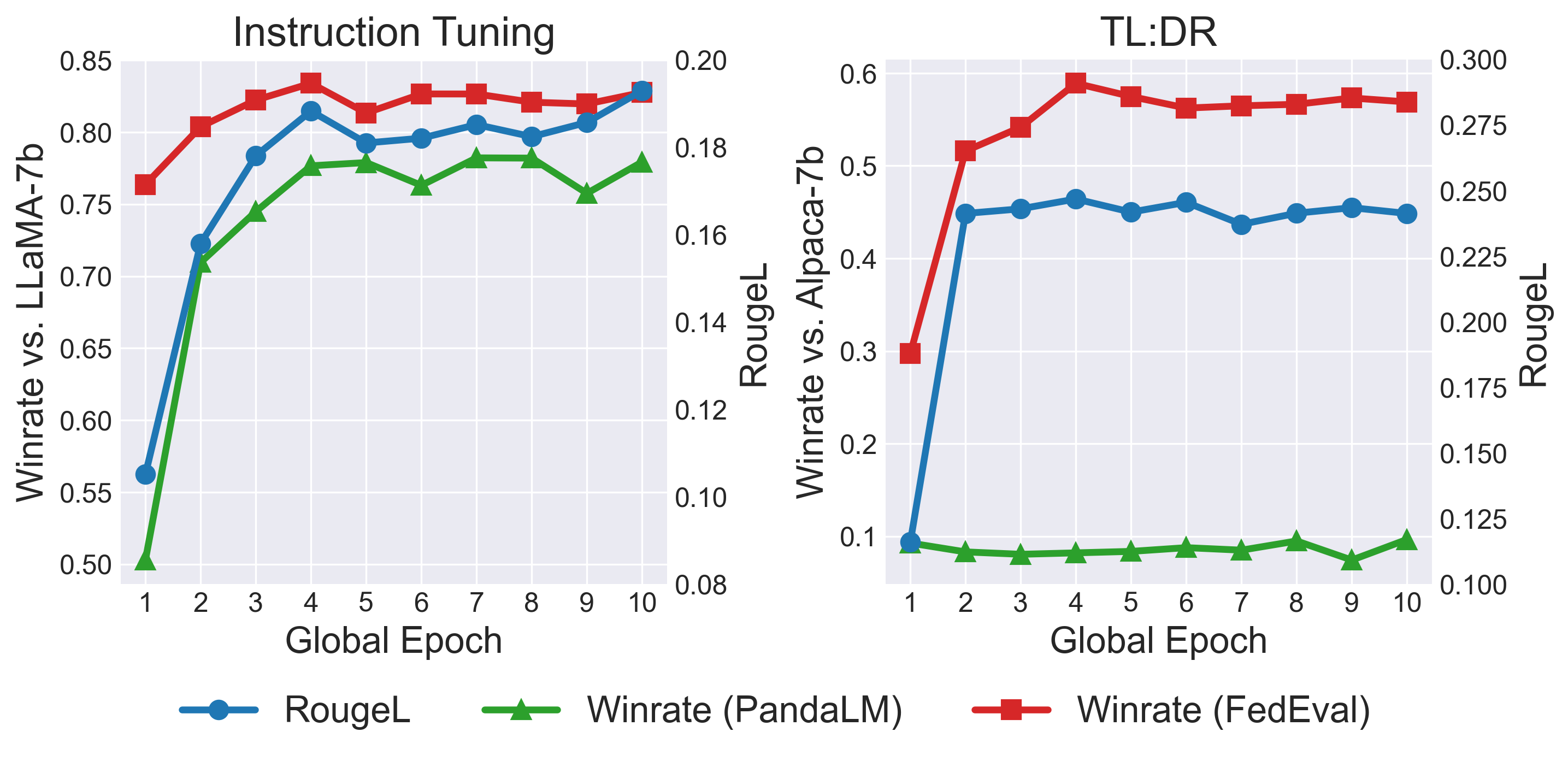}\\
\vspace{3mm}
\parbox[c]{8.3cm}{\footnotesize{Fig.3.~}  Performance of aggregated global model measured by RougeL-score and win-rate against LLaMA-7B model evaluated by PandaLM and three personalized evaluation models.}
\label{fig:eval_global}
\end{center}

We measure the performance of local models as the win-rate against the base model, LLaMA-7B for instruction tuning task and Alpaca-7B\jz{1}{https://github.com/tatsu-lab/stanford$\_$alpaca} for TL;DR summary task. For the instruction tuning task, the result obtained from PandaLM and the RougeL-score exhibit a similar trend. Both indicate that the performance of the global model mainly improves during the first 4 epochs, followed by oscillations. Notably, the FedEval-LLM result is in good agreement with these two reference results, demonstrating its competitive evaluation capability. For the TL;DR summary task, PandaLM consistently favors the result of the base model (low win rate) and thus fails to reflect true model performance. Upon closer examination of individual results, we identified that this preference arises from the base model's tendency to produce longer summaries, which is favored by PandaLM. In contrast, FedEval-LLM's results strongly correlate with the RougeL-score. This again demonstrates that domain-specific knowledge is necessary for accurate evaluation. As a result, the proposed FedEval-LLM framework is capable of efficiently utilizing this knowledge and providing reliable evaluations.

\noindent{\textbf{Local Model Evaluation.}} In addition to assessing the global model's performance, it is also desired to discern the performance of local models, particularly when determining aggregation weights or contributions. This is more challenging compared to the evaluation of the global model due to the potential subtle difference among local models. The results are presented in Fig. 4.

\setcounter{figure}{3}
\begin{center}
\includegraphics[width=8.5cm]{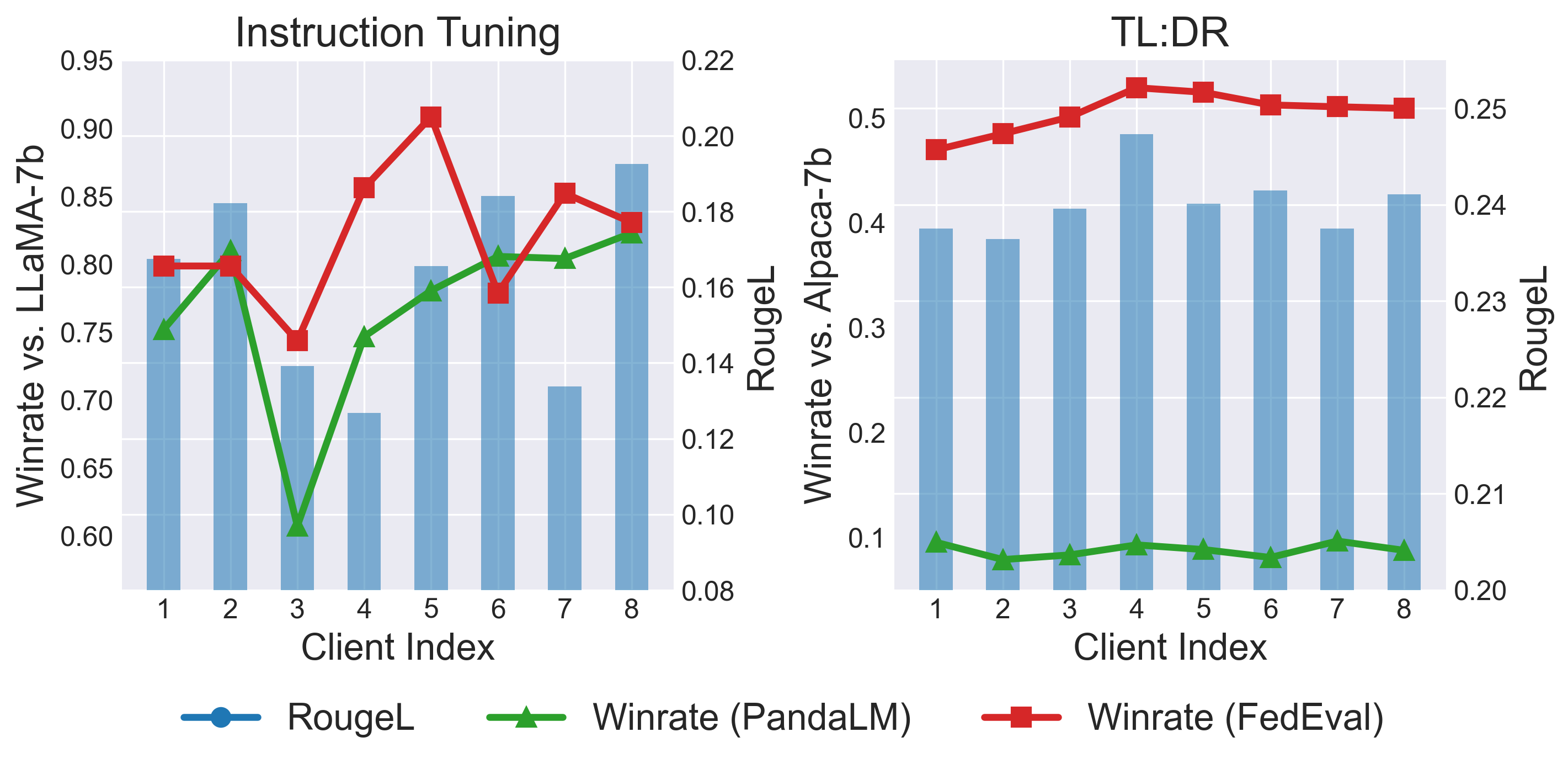}\\
\vspace{3mm}
\parbox[c]{8.5cm}{\footnotesize{Fig.4.~} Performance of local model measured by RougeL-score and win-rate against Alpaca-7B model evaluated by PandaLM and three personalized evaluation models.}
\label{fig:eval_local}
\end{center}

For the instruction tuning task, the results of PandaLM and RougeL largely follow a similar trend. However, they exhibit noticeable deviations for clients 3, 4, and 7. In contrast, FedEval-LLM aligns more closely with PandaLM results than with RougeL-score, indicating its capability to capture the nuanced performance difference among local LLMs. This observation also suggests that the similarity-based metric may not be optimal for open-ended generative tasks. For the TL;DR summary task, variations among the clients are considerably smaller compared to the instruction tuning task, mainly due to more confined task space. Therefore, it is expected that the RougeL-score could more faithfully reflect the performance for this kind of task. While PandaLM could not provide meaningful results owing to its lack of domain knowledge, similar to the case of global model evaluation, FedEval-LLM demonstrates good agreement with RougeL-score, with both peaking at client 4. This clearly suggests that the personalized evaluation model, trained on the evaluation data of the target task, better aligns with the target evaluation criteria and is more suitable than the general-purpose evaluation model that lacks knowledge of the target task. 

\section{Discussion}

\subsection{The Effectiveness of Selection Criteria}
To demonstrate the effectiveness of the proposed three selection criteria, we measure the evaluation capability of trained evaluation models on the instruction tuning task with limited available evaluation data, taking into consideration the constraints posed by communication and computation overhead. We employ the adaptive bootstrapping strategy described in Section 4.2.1, where valid evaluation samples from the first round are utilized to train temporary evaluation models that, in turn, are used for subsequent generation rounds. The results are presented in Fig. 5.

\setcounter{figure}{3}
\begin{center}
\includegraphics[width=8.5cm]{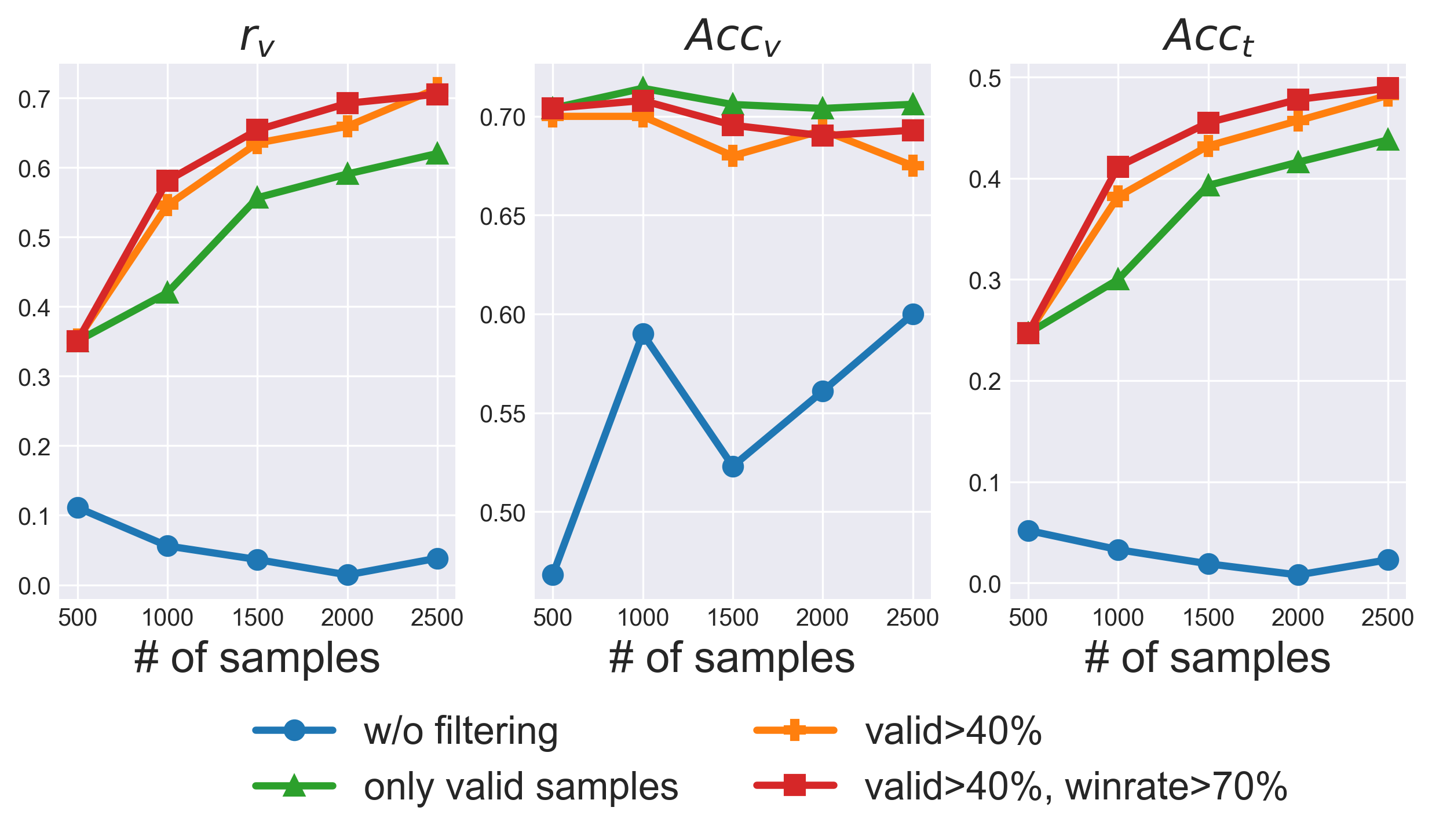}\\
\vspace{3mm}
\parbox[c]{8.5cm}{\footnotesize{Fig.5.~} Performance of personalized evaluation models trained with varying number of training samples on instruction tuning task. The results are obtained by averaging 8 clients. }
\label{fig:ft_vs_sample}
\end{center}

It is evident that when none of the criteria are employed, both $r_v$ and $Acc_v$ remain significantly low, confirming the limited evaluation capability of local LLM and, as a result, the low quality of raw evaluation data. By solely using the order-consistency criterion, we observe a notable improvement in $r_v$, surpassing $0.3$, and $Acc_v$ close to $0.7$. This demonstrates the effectiveness of the order-consistency criterion in eliminating noisy data. Furthermore, we note that while $r_v$ increases with more training samples, $Acc_v$ remains consistently around $0.7$. By applying the other two criteria, we observe a further enhancement of $r_v$. This suggests that these criteria effectively identify more consistent evaluation samples, thus enhancing the intrinsic evaluation capabilities. Consequently, when all three criteria are employed, we obtain the best evaluation models, as indicated by the highest $Acc_t$ (represented by the red line). 

\subsection{The Importance of Knowledge on Downstream Tasks}

As demonstrated in our prior experiments, PandaLM~\cite{wang2023pandalm}, trained on over 300K evaluation data from the instruction tuning task, exhibits limited evaluation capability on the TL;DR summary task. In contrast, FedEval-LLM demonstrates the ability to provide reliable evaluations by leveraging the domain knowledge of each client. This observation underscores the importance of domain knowledge for the corresponding evaluation capability. To gain deeper insights into the factors affecting evaluation capabilities in downstream tasks, we conducted a preliminary experiment using the TL;DR dataset, employing varying quantities and combinations of standard training data and evaluation data. The corresponding results are presented in Table 5.

It is evident that training solely on instruction tuning data,  without fine-tuning on evaluation data, yields minimal benefits to the evaluation capability of the TL;DR summary task, as exhibited by the low $r_v$ and $Acc_v$ value. Training on TL;DR data leads to marked improvements in both $r_v$ and $Acc_v$, suggesting domain knowledge is essential. Although $Acc_v$ shows significant improvement as more training data is used, increasing from 0.004 to 0.087, $r_v$ declines from 0.759 to 0.310. As a result, it leads to the enhanced overall evaluation capability but is still limited, as exhibited by $Acc_t$. We think that the increased quantity of training data enhances the model's capability for instruction following (improved $r_v$). However, it is less effective in enhancing the task-related evaluation capability. 

Fine-tuning on evaluation data yields a substantial increase in $r_v$, suggesting an improved inherent evaluation capability. However, incorporating evaluation data from the instruction tuning task brings limited improvement to $Acc_v$ and even reduces the valid accuracy when 12K TL;DR data are used. This suggests that the evaluation criterion employed in the instruction tuning task is not in alignment with the TL;DR summary task, i.e., it lacks task-related evaluation capability. In contrast, integrating evaluation data from the TL;DR summary task enhances both $r_v$ and $Acc_v$, demonstrating the evaluation data on the target task is imperative for gaining task-related evaluation capability. In summary, we obtain two observations: 1) Training on data from the target domain can enhance evaluation capability on the target task, and 2) In-domain evaluation data is indispensable for gaining task-specific evaluation capabilities, while out-of-domain evaluation data can enhance the model's inherent evaluation capability, but offers minimal improvement in task-specific evaluation capabilities.

\tabcolsep 10pt
\renewcommand\arraystretch{1.3}
\begin{center}
{\footnotesize{\bf Table 5.} Evaluation Capability of fine-tuned LLM on TL;DR dataset.}\\
\vspace{2mm}
\footnotesize{
\begin{tabular*}{\linewidth}{l|c|c|c|c}\hline\hline\hline
        \hline
		Local Data & Eval Data & $r_v$ & $Acc_v$ & $Acc_t$ \\ 
	    \hline
        \hline
        \multirow{3}{*}{8k IT} & - & 0.000 & /  & / \\
        ~ & 800 IT & 0.272 & 0.445 & 0.121\\
        ~ & 800 TL;DR & 0.432 & 0.634 & 0.274\\
        \hline
        \multirow{3}{*}{4k TL;DR} & - & 0.004 & 0.750 & 0.003 \\
        ~ & 800 IT & 0.075 & 0.780 & 0.058\\
        ~ & 800 TL;DR & \textbf{0.091} & \textbf{0.827} & \textbf{0.075}\\
        \cline{2-5}
		\multirow{3}{*}{8k TL;DR}  & - & 0.017 & 0.471 & 0.008 \\
        ~ & 800 IT & 0.313 & 0.540 & 0.169\\
        ~ & 800 TL;DR & \textbf{0.577} & \textbf{0.598} & \textbf{0.345}\\
		\cline{2-5}
        \multirow{3}{*}{12k TL;DR}  & - & 0.087 & 0.310 & 0.027\\
         ~ &  800 IT & 0.565 & 0.149 & 0.084 \\
         ~ & 800 TL;DR & \textbf{0.697} & \textbf{0.650} & \textbf{0.453} \\
\hline\hline\hline
\end{tabular*}
\\\vspace{1mm}\parbox{8.3cm}{TL;DR evaluation data is extracted from the test set of ~\cite{stiennon2022learning}. Local data is sampled from Dolly15K and TL;DR summary datasets, respectively. Eval data is evaluation data from~\cite{zheng2023judging} and training set from ~\cite{stiennon2022learning}, respectively. Here, \textit{IT} stands for instruction tuning data.}
}
\label{table:ft_summary}
\end{center}

\subsection{Privacy}
Privacy preservation is the fundamental prerequisite in Federated Learning (FL). However, the privacy concerns associated with FL combined with LLMs have not been extensively examined~\cite{fan2023fatellm}. The proposed FedEval-LLM framework addresses two primary apprehensions regarding the potential data leakage risks in the evaluation: 1) FedEvaL-LLM utilizes evaluation models from participants to provide reliable evaluation, thus obviating the need for external servers and mitigating the risk of data leakage to third parties. 2) FedEval-LLM exclusively mandates test sets comprising questions for both constructing the evaluation model and deploying it for assessment, thereby further minimizing the risk of leaking information related to local private data.

We further delve into the privacy implications inherent in our proposed FedEval-LLM framework, considering a prevalent threat model in FL where participants and the server are semi-honest~\cite{lyu2020threats}. In this context, they strictly follow the predefined instructions, but may attempt to infer the private information from local data of other participants. This information could encompass the raw local data itself or more abstract knowledge, such as details regarding the domain and distribution of the data.

First, FedEval-LLM eliminates the necessity for labeled test sets. The questions from the test set used for training the evaluation model and for the collective evaluation can be sourced from publicly available repositories or collaboratively derived through a consensus among participants. This alleviates the potential risk of data leakage associated with labels.

Second, the information exchanged during the training of evaluation models and the evaluation process is restricted to evaluation results and the responses to the test set. The model parameters of both local models and evaluation models are not shared. The evaluation results merely expose model preference. As for the answers, constraining the scope of questions from the test set allows effective control over the corresponding responses of LLMs, thereby preventing the potential leakage of information regarding the domain or distribution of local data. Since model parameters are not shared, attacks that aim to infer raw text using gradients of language models are not viable~\cite{balunovic2022lamp,gupta2022recovering}.

In light of the preceding discussion, we posit that FedEvaL-LLM exhibits robust privacy-preserving capabilities by confining the information exchange process to FL participants and the scope of exchanged information within the realm of predefined questions.

\section{Conclusion}
In this work, we propose a federated evaluation framework, FedEval-LLM, that can provide reliable evaluations of LLM's performance on downstream tasks while preserving the privacy of local data. It promotes the evaluation capability of participants by training a personalized evaluation model using local data and bootstrapped task-specific evaluation data. We experimentally demonstrated that this approach can significantly enhance the LLM's evaluation capability on the target task. By collectively employing the evaluation model, we can accurately monitor the performance of the global model, as well as the minor performance variations among local models. Notably, in these two stages, the employment of a group of LLMs as referees plays a pivotal role in mitigating noisy and biased evaluations. We demonstrated that domain knowledge is critical for gaining corresponding evaluation capability, which can only be harnessed through training on local data. Furthermore, we discussed the potential privacy risks associated with the evaluation under FL, and we assert that FedEval-LLM provides strong privacy-preserving capability compared to other methods by obviating the need for external services and labeled test sets. We hope that the proposed paradigm, which leverages LLM's multi-task capability via training personalized model, could inspire future works that seek to integrate LLM within the FL framework.

\bibliographystyle{JCST}
\bibliography{ssl_vfl}

\end{multicols}
\label{last-page}
\end{document}